\documentclass[10pt,twocolumn,letterpaper]{article}

\usepackage{tabularx,colortbl}
\usepackage[pagenumbers]{cvpr} 

\usepackage[dvipsnames]{xcolor}

\usepackage{amsmath,amsfonts}

\DeclareMathOperator*{\argmin}{arg\,min} 

\definecolor{cvprblue}{rgb}{0.21,0.49,0.74}

\usepackage{hyperref}
\hypersetup{pagebackref,breaklinks,colorlinks,citecolor=cvprblue}

\def\paperID{16714}
\def\confName{CVPR}
\def\confYear{2024}

\title{A Quantitative Evaluation of Score Distillation Sampling Based Text-to-3D}

\author{Xiaohan Fei${}^*$ \qquad Chethan Parameshwara\thanks{Equal contribution.} \qquad Jiawei Mo \qquad Xiaolong Li\\ \qquad Ashwin Swaminathan \qquad CJ Taylor \qquad Paolo Favaro \qquad Stefano Soatto\\AWS AI Labs\\{\tt\small \{xiaohfei, cmparam\}@amazon.com}}

\begin{document}
\maketitle

\begin{abstract}
The development of generative models that create 3D content from a text prompt has made considerable strides thanks to the use of the score distillation sampling (SDS) method on pre-trained diffusion models for image generation. However, the SDS method is also the source of several artifacts, such as the Janus problem, the misalignment between the text prompt and the generated 3D model, and 3D model inaccuracies. While existing methods heavily rely on the qualitative assessment of these artifacts through visual inspection of a limited set of samples, in this work we propose more objective quantitative evaluation metrics, which we cross-validate via human ratings, and show analysis of the failure cases of the SDS technique. We demonstrate the effectiveness of this analysis by designing a novel computationally efficient baseline model that achieves state-of-the-art performance on the proposed metrics while addressing all the above-mentioned artifacts.
\end{abstract}

\section{Introduction}
\label{sect-intro}

One of the most appealing aspects of text-conditioned generative models is that they allow to create assets, such as images, videos and 3D models, that would only be possible by turning to highly-skilled experts. While the generation of images has now reached a significant level of sophistication, the creation of 3D models from text is still at its infancy despite the tremendous progress in the last year. In this paper, we thus focus on this family of models.

One of the main open challenges in designing and improving these generative models is that there is no systematic quantitative evaluation protocol in the literature. Most work only provides qualitative visual comparisons or user studies against competitors. Thus, such evaluations do not provide a comprehensive quantitative feedback on how well models are working, nor give any guidance on how to avoid failures. For instance, we observe that in many state-of-the-art methods the generated 3D content often presents artifacts such as the ``Janus problem'' (as reported in Table~\ref{tab-janus}), where the reconstructed 3D model consists of multiple (incorrectly) repeated object parts (see examples in Figure~\ref{fig-janus}). The Janus problem has been pointed out by a few authors~\cite{poole2023dreamfusion,shi2023mvdream,chen2023text-gsgen,he2023t3bench}, but it has never been measured quantitatively. Therefore, we propose a novel metric to measure the severity of the Janus problem as part of our evaluation protocol. In addition, we also measure the alignment between the text prompts and the generated 3D content as well as the realism of the generated 3D models (see section~\ref{sect-metrics}).

To illustrate the effectiveness of our proposed metrics, we also present a novel baseline method for text-to-3D, which achieves state-of-the-art performance. When choosing the framework for our approach, we notice that there are two main paradigms in the text-to-3D literature. The first paradigm aims to directly learn the posterior distribution of 3D models given a prompt by utilizing large repositories of CAD models (\eg, Objaverse~\cite{deitke2023objaverse}) paired with text descriptions.
Approaches within this paradigm include Point-E~\cite{nichol2022pointe} and Shap-E~\cite{jun2023shape}. Although more and more large-scale 3D datasets such as Objaverse are being made available, they are not yet a match, in scale, to their 2D counterparts (\eg, LAION~\cite{schuhmann2022laion}). Because of this relative scarcity of 3D data, models built in this paradigm have limited generalization capabilities. Moreover, because these datasets are synthetic, there is a fundamental challenge in eliminating the domain gap with real data. 

The second paradigm in text-to-3D leverages text-to-image~\cite{rombach2022high-ldm} models pre-trained on Internet-scale image data (\eg, LAION~\cite{schuhmann2022laion}). 
The key idea is to build a 3D representation by exploiting the multiview information stored into a pre-trained text-to-image model. Such models provide a direct feedback to the neural renderings of views of a 3D representation through a technique introduced by Poole~\etal~\cite{poole2023dreamfusion} called Score 
Distillation Sampling (SDS) and later adopted by many others.
Currently, methods in the second paradigm yield the most appealing results: Their generative models can handle more complex text prompts and provide more realistic-looking, diverse and highly-detailed 3D content than methods in the first paradigm. This motivates us to consider the SDS approach in our work.

In addition to the problems mentioned above, a limitation of SDS approaches for their applicability to real-world use cases is that they require a time-consuming optimization procedure at inference time. We propose to use Gaussian Splatting~\cite{kerbl20233d-gaussiansplatting} as a building block to accelerate the process. Using Gaussian Splatting in text-to-3D has recently been explored by \cite{tang2023dreamgaussian,chen2023text-gsgen,yi2023gaussiandreamer}. Compared to these recent developments, we show that our approach achieves competitive inference efficiency (Table~\ref{tab-efficiency}), while outperforming them in terms of the frequency of the Janus problem (Table~\ref{tab-janus}), the text and 3D alignment (Table~\ref{tab-alignment}), and the realism of the generated models (Table~\ref{tab-fidelity} and Figure~\ref{fig-qualitative-comparison}). Compared to methods {\em not} based on Gaussian Splatting, we achieve at least {\em twice} a speed up, while being competitive in the other metrics. We expect that our proposed evaluation protocol will move the development of text-to-3D methods forward.

To summarize, our contribution is threefold: i) We propose an evaluation protocol to thoroughly and objectively evaluate the performance of text-to-3D methods covering the Janus problem, text and 3D alignment, and the realism of the generated 3D models (see section~\ref{sect-protocol}); ii) we analyze the SDS framework, its pitfalls -- and, in particular, the ``Janus problem'', their causes, and possible mitigation (see section~\ref{sect-formulation}); iii) we propose a new baseline text-to-3D method by incorporating the latest developments in the field (see section~\ref{sect-method}). Thanks to our proposed evaluation protocol, we are able to properly evaluate our method, and show that it is both efficient and competitive in all the proposed metrics when compared to the state of the art.

\section{Problem formulation and analysis}
\label{sect-formulation}
We  first formalize the SDS framework in section~\ref{sect-sds}, and then conduct analysis in section~\ref{sect-sds-issues}.

\subsection{Score distillation sampling}
\label{sect-sds}
Common choices for a 3D representation are models with parameters $\theta$ such as NeRF~\cite{mildenhall2021nerf}, SDF~\cite{yariv2021volume}, Mesh~\cite{shen2021deep-dmtet}, and Gaussian Splatting~\cite{kerbl20233d-gaussiansplatting}. We adopt Gaussian Splatting in our method for its efficiency, but the choice of the parametrization will not affect our analysis.
Let $g_\theta(\cdot)$ be a differentiable rendering function that takes a camera pose $c\in\mathrm{SE}(3)$ as input and outputs an image $x=g_\theta(c)$. Let $y$ denote the text prompt given by a user. Then, a Maximum Likelihood objective to realize text-to-3D is
\begin{equation}
\argmin_\theta \mathbb{E}_c[ - \log p(x | y, c)].
\label{eq-log-likelihood-objective}
\end{equation}
To perform gradient descent on this objective, one needs to compute the gradient $\mathbb{E}_c[-\nabla_\theta \log p(x|y,c)]$. The term in the square brackets can be computed through the chain rule as the product of the score function and the gradient of the rendering function, \ie, $\nabla_\theta \log p(x|y, c)=\nabla_x \log p(x|y, c)\frac{\partial g_\theta(c)}{\partial \theta}$. As we explain below, the score function $\nabla_x \log p(x|y, c)$ can be approximated by using a pre-trained and frozen text-to-image diffusion model. However, because these models are not conditioned on the camera pose $c$, one common practice in the literature is to ``prompt engineer'' a pre-trained and frozen text-to-image model with viewpoint phrases. Specifically, the camera configuration $c$ is first quantized into a viewpoint phrase (\eg, ``front view'') which is then appended/prepended to the original prompt $y$ (\eg, ``a corgi'') resulting in an augmented prompt $\tilde y$ (\eg, ``a corgi, front view'') which is finally used in the approximation $\nabla_x \log p(x|y, c) \approx \nabla_x \log p(x|\tilde y)$.
With this approximation, Eq.~\eqref{eq-log-likelihood-objective} can be approximated as  $\argmin_\theta \mathbb{E}_c[ - \log p(x| \tilde y)]$.
The above optimization is a hard one, especially when one starts from a randomly initialized 3D representation which renders images of pure noise at the onset of training. To reconcile this challenge, as commonly done in the literature, the objective is optimized across multiple noise levels corresponding to different time steps $t\in 1\cdots T$ leading to what is called the score distillation sampling (SDS) loss below
\begin{equation}
\argmin_\theta \mathbb{E}_{c, t} [ - \log p_t(x_t| \tilde y)]
\label{eq-sds-loss}
\end{equation}
where $x_t$ is a noisy version of $x$. Depending on the implementation details and output type (noise, sample, or velocity) of the text-to-image's UNet component, the objective could take slightly different forms. For instance, a noise prediction UNet is often used in the literature -- which we also adopt in our implementation. However, in our analysis below, we will use the score function formulation as it abstracts away the underlying implementation details, is mathematically equivalent to other forms, and makes the analysis more approachable.

\subsection{Analysis of score distillation sampling}
\label{sect-sds-issues}
\begin{figure}[t]
\centering
\includegraphics[width=1\linewidth]{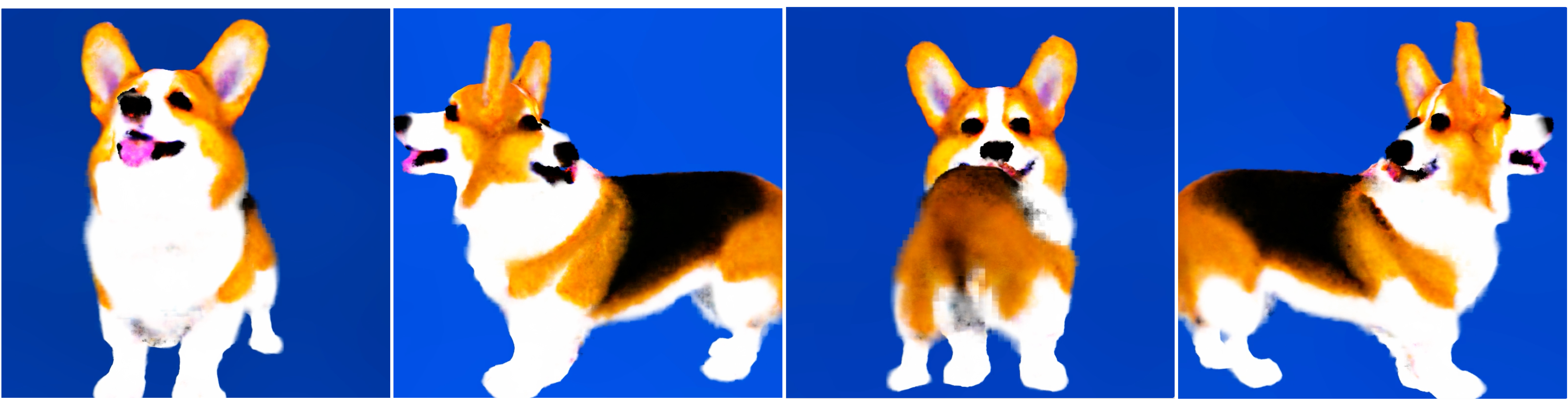}
\caption{\textit{Demonstration of the Janus problem} in DreamFusion~\cite{poole2023dreamfusion} -- a state-of-the-art text-to-3D model. Prompt: ``A corgi.'' We show renderings of the generated 3D corgi from different viewpoints (left to right): Front, left, back, and right view of the corgi.}
\label{fig-janus}
\vspace{-0.2in}
\end{figure}
In this section, we analyze the objective in Eq.~\eqref{eq-log-likelihood-objective} and its approximation in Eq.~\eqref{eq-sds-loss}.
(i) First, for Eq.~\eqref{eq-log-likelihood-objective} to work, one needs to define a coordinate frame in which the camera pose $c$ can be defined. However, the choice of the coordinate frame is arbitrary. To see this, consider attaching a coordinate frame to a rigid object. Should the origin of the coordinate frame coincide with the center of the object, or should it be placed on the surface supporting the object? And how should the coordinate frame be oriented?
(ii) Second, even if there is a shared agreement on a canonical coordinate system for each possible 3D scene in the world, text-to-image models that are utilized by SDS are not trained with viewpoint conditioning.
(iii) Third, given a text prompt, \eg, ``a corgi'', text-to-image models are trained to generate samples of {\em possibly different corgi instances under nuisance variabilities including varying background, illumination and deformation}. Therefore, to obtain a point estimate of ``a corgi'' in text-to-3D, one needs to marginalize out all these nuisance variabilities which is a challenge.~\footnote{Most work try to marginalize the nuisance variabilities in training through data augmentation (\eg, randomize the lighting condition and camera focal length). ProlificDreamer~\cite{wang2023prolificdreamer} instead seeks a distribution instead of a point estimate of possible 3D models. But these techniques are insufficient to account for all nuisance variabilities such as deformation.}

Given all the challenges above, it seems impossible to use SDS to conduct the text-to-3D task. Yet, DreamFusion~\cite{poole2023dreamfusion} and follow-up papers have shown that the idea could work -- though we found experimentally that these approaches do not always output a valid 3D model. We analyze why this is the case in the following three points:
(i) First, even though the choice of the coordinate frame is arbitrary, we hypothesize that, statistically, most people can agree on some approximate definition of the canonical frame. This could be captured by the large amount of text-image pairs available on the Internet, which are used to train the text-to-image models. For instance, when prompted with ``a corgi, front view'', we observe that the majority of images generated by Stable Diffusion~\cite{rombach2022high-ldm} contains a corgi looking at the camera, which corroborates our hypothesis. (ii) This means that even though text-to-image models such as Stable Diffusion are not explicitly conditioned on the {\em exact camera viewpoints}, they are able to condition on the {\em viewpoint phrases} that approximate the exact viewpoints. But since the viewpoint phrases are just approximations, we found that conditioning on them could fail from time to time leading to the Janus problem as seen in Figure~\ref{fig-janus}. (iii) To reconcile the third challenge above, a high classifier-free guidance (CFG) scale is often used in the literature to encourage the mode seeking behavior of text conditioned diffusion models at the cost of reduced diversity in the generated content. Yet, a high CFG scale, though effective, does not guarantee to eliminate all the nuisance variabilities, and as such, the possible deformation of the object (among many other variabilities) could also lead to erroneous geometry, for example, the Janus problem as shown in Figure~\ref{fig-janus}.

Our analysis is also supported by our experiments (section~\ref{sect-protocol}) where we found that most of the state-of-the-art methods fail very often (more than half of the time) even on very simple prompts (\eg, ``a corgi'') due to the Janus problem and thus rendering these methods not yet applicable in real-world use cases. To properly assess text-to-3D methods, we advocate to measure the severity of the Janus problem as part of the evaluation protocol detailed in section~\ref{sect-protocol}. Equipped with a proper evaluation protocol, we draw inspirations from various state-of-the-art methods, and propose our own method in section~\ref{sect-method}, thus setting a strong baseline for future text-to-3D models.

\section{Method}
\label{sect-method}
We first describe the two building blocks of our method: MVDream~\cite{shi2023mvdream} and Gaussian Splatting~\cite{kerbl20233d-gaussiansplatting}.
Then, we describe our two-stage method that is efficient, (almost) Janus-free, and generates realistic 3D content.

\subsection{Multiview diffusion}
\label{sect-mvdream}
MVDream~\cite{shi2023mvdream} proposes to learn the score function of multiple views of the same scene jointly -- that is $\nabla_{x_{1:N}} \log p(x_{1:N} | y, c_{1:N})$, where $x_{1:N}$ indicates $N$ views of the same 3-D scene and $c_{1:N}$ indicates their corresponding camera poses.
MVDream builds on Stable Diffusion. To model the joint likelihood of multiple views, a cross attention mechanism is introduced throughout Stable Diffusion's UNet, where the latents of $N$ views are attended jointly. The model can thus be trained on a varying number of views including a single view as a special case. In practice, MVDream is trained on both multiview renderings of the Objaverse~\cite{deitke2023objaverse} 3D dataset as well as LAION~\cite{schuhmann2022laion} to improve the realism of its generation.
To use MVDream in the SDS framework, the objective below is optimized
\begin{equation}
\argmin_\theta \mathbb{E}_{c_{1:N}}[-\log p(x_{1:N} | y, c_{1:N})]
\label{eq-sds-mvdream}
\end{equation}
where $x_{1:N} \doteq g_\theta(c_{1:N})$. This objective is also optimized across multiple noise levels indexed by the time step $t$. We drop $t$ to unclutter the notation. Unlike the objective in Eq.~\eqref{eq-sds-loss}, here we use the exact camera poses defined in the coordinate frame of the CAD models, which helps to reduce the Janus problem as shown in our experiments in section~\ref{sect-protocol}.

\subsection{Gaussian Splatting}
\label{sect-gs}
Gaussian Splatting~\cite{kerbl20233d-gaussiansplatting} is a recent neural scene representation that enables efficient learning and {\em real-time rendering} of high-resolution videos of 3D scenes. Unlike the standard NeRF~\cite{mildenhall2021nerf} or its accelerated variants~\cite{muller2022instant}, Gaussian Splatting is an explicit representation where the scene is modeled by a collection of Gaussian balls. Each Gaussian ball is associated with its position, covariance, opacity, scale, and view-dependent appearance parameters (\ie, coefficients of Spherical Harmonics) that could be learned. An efficient differentiable rasterization method is also introduced in \cite{kerbl20233d-gaussiansplatting} enabling end-to-end learning of these parameters.

\subsection{Proposed baseline method}
\label{sect-proposed-method}
Our baseline method consists of two stages.
In the first stage, we optimize the objective in Eq.~\eqref{eq-sds-mvdream}, where a pre-trained multiview-aware diffusion model (\ie, MVDream) is used to obtain a coarse, but multiview consistent, 3D model.
The second stage refines the output of the first stage with a joint loss (Eq.~\eqref{eq-sds-loss} and Eq.~\eqref{eq-sds-mvdream}) based on two separate diffusion models that encourages fine-grained details with added realism, while preserving the multiview consistent 3D geometry as much as possible.

\begin{figure}[t]
\centering
\includegraphics[width=0.85\linewidth]{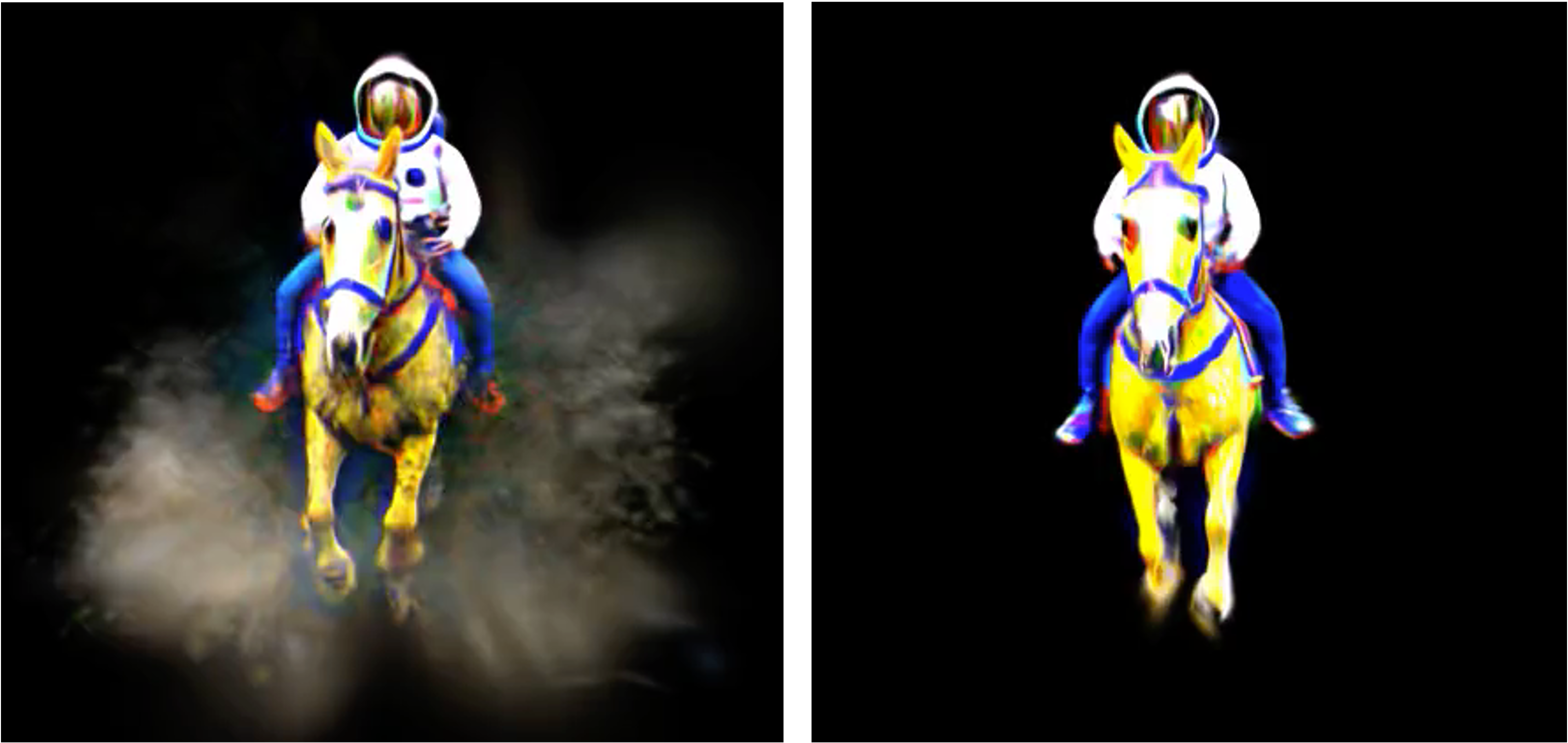}
\caption{\textit{Ablation of the regularization term.} Prompt: ``An astronaut riding a horse.'' Left: our method without any regularization. Right: our method with the regularization term Eq.~\eqref{eq-sparsity}, which effectively reduces floaters.}
\label{fig-regularization}
\end{figure}

Gaussian Splatting is used in both stages. Unlike NeRF which consists of MLPs and as such has smoothness properties by construction (\eg, layers in an MLP, such as the ReLU, are piecewise linear), Gaussian Splatting consists of a set of independent Gaussian balls and thus does not automatically have such properties. Thus, a vanilla Gaussian Splatting implementation of text-to-3D would suffer from significant artifacts, such as \emph{floaters} (\ie, small spurious volume elements in the space between the object and the camera views).
To reduce these artifacts, we introduce a regularization term on the rendered \emph{alpha maps} (\ie, the silhouette of the object projected onto a camera view)
\begin{equation}
    L_{\text{sparsity}} = \mathbb{E}_c[ \frac{1}{|\Omega|} \| \alpha_\theta(c) \|_1 ]
    \label{eq-sparsity}
\end{equation}
where $\Omega\subset\mathbb{R}^2$ denotes the image domain, and $\alpha_\theta(\cdot)$ refers to the differentiable rendering function that takes the camera pose $c$ as input and outputs the alpha map. The subscript of $\alpha_\theta$ indicates that the rendered alpha map is a function of the model parameters $\theta$ which, in our case, consists of the Gaussians' attributes as described in section~\ref{sect-gs}.
Figure~\ref{fig-regularization} showcases the effectiveness of our regularization.

\begin{figure}[t]
\centering
\includegraphics[width=1.0\linewidth]{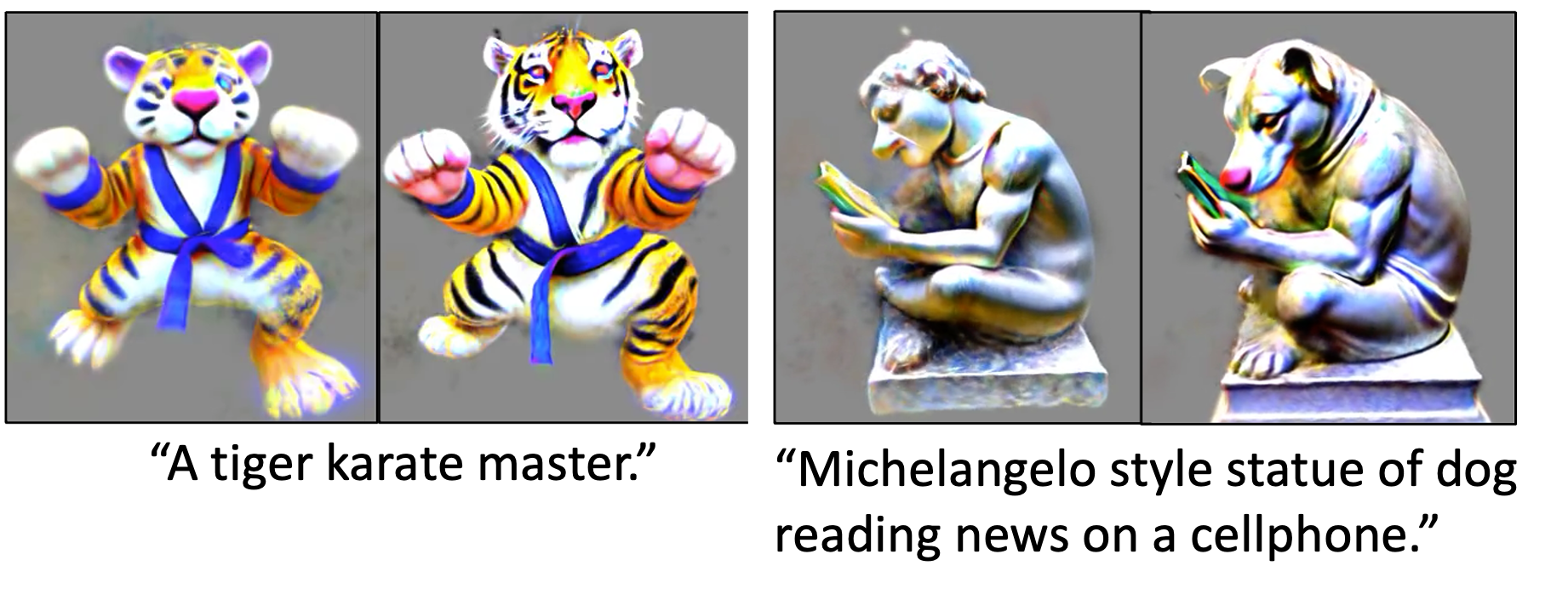}
\vspace{-0.2in}
\caption{\textit{Ablation of the refinement stage.} Two examples are shown. In each example, the image on the left shows the rendering of our first stage model, and the one on the right shows the rendering of our full model. The refinement stage greatly improves the alignment between the 3D content and the text prompt as well as the realism of the 3D content.}
\label{fig-secondstage}
\vspace{-0.1in}
\end{figure}

Since the multiview diffusion model used in the first stage is trained on the Objaverse 3D dataset, we observe that the generated content often has limited diversity and lacks realism (see Figure~\ref{fig-secondstage}). To correct them, we introduce a refinement stage.
In this stage, we refine the Gaussian parameters obtained in the first stage by minimizing a joint loss which is a weighted sum of Eq.~\eqref{eq-sds-loss} and Eq.~\eqref{eq-sds-mvdream} along with the regularization term (Eq.~\eqref{eq-sparsity}). Eq.~\eqref{eq-sds-mvdream} is realized with the same pre-trained MVDream model as in the first stage. To realize Eq.~\eqref{eq-sds-loss}, we use a Stable Diffusion model pre-trained only on real data. The use of Stable Diffusion in the form of an additional SDS loss (\ie, through Eq.~\ref{eq-sds-loss}) in our refinement stage is fundamental to close the domain gap of MVDream and to improve the diversity and quality of the generated 3D models (see Figure~\ref{fig-secondstage}). However, as analyzed in section~\ref{sect-sds-issues}, the use of Eq.~\ref{eq-sds-loss} and text-to-image models that are not trained to condition on viewpoints (\eg, Stable Diffusion) are at the risk of causing the Janus problem, which we also observe in our refinement stage. This can be seen in Table~\ref{tab-janus} and Table~\ref{tab-fidelity} where our refinement stage improves the fidelity of the generated 3D content, but in the same time introduces more Janus problems. That said, a tradeoff needs to be made between the Janus problem and the fidelity of the generated content. We study this tradeoff in the supplementary material.

\noindent\textbf{Implementation details} We build our model atop the open-source ThreeStudio~\cite{threestudio2023} framework. We train the first stage for 2000 steps followed by 1000 steps in the refinement stage. We start with 5000 randomly initialized Gaussian balls, and perform densification every 100 steps until half the number of iterations and pruning every 100 steps until the end of the iterations. Throughout the training, the weight for the MVDream SDS loss (Eq.~\eqref{eq-sds-mvdream}) and the regularization term (Eq.~\eqref{eq-sparsity}) are both set to 1. In the second stage, the weight for the Stable Diffusion SDS loss (Eq.~\eqref{eq-sds-loss}) is 0.5.
A CFG scale of 50 is used in both MVDream and Stable Diffusion for text conditioning.
More implementational details could be found in the supplementary material. 

\section{Proposed evaluation protocol}
\label{sect-protocol}
Most of the prior work in text-to-3D only conducts qualitative visual comparisons between competing methods. However, a quantitative characterization of the performance of each method is necessary to drive the progress forward. We provide a more thorough and systematic evaluation protocol in this section. We first describe the state-of-the-art methods evaluated in our experiments (section~\ref{sect-baselines}) followed by which prompts we use (section~\ref{sect-prompts}).
Then, in section~\ref{sect-metrics}, we describe our quantitative metrics along with the comparison results.
We also provide an efficiency metric, which measures the GPU-hours needed by each method to generate one 3D model end-to-end (section~\ref{sect-efficiency}).

\subsection{Considered state-of-the-art methods}
\label{sect-baselines}
\noindent\textbf{DreamFusion}~\cite{poole2023dreamfusion} introduces the SDS framework and advocates the idea of using pre-trained text-to-image diffusion models for text-to-3D.\\
\noindent\textbf{DreamFusion+PerpNeg}~\cite{armandpour2023perpneg} is built on DreamFusion and proposes a technique to better condition text-to-image models on viewpoint phrases to mitigate the Janus problem.\\
\noindent\textbf{Magic3D}~\cite{lin2023magic3d} proposes a two-stage method, where the first stage is similar to DreamFusion, but uses an efficient NeRF variant (\ie, Instant-NGP~\cite{muller2022instant}), and the second stage refines a differentiable mesh representation (DMTet~\cite{shen2021deep-dmtet}) extracted from the first stage's NeRF.\\
\noindent\textbf{TextMesh}~\cite{tsalicoglou2023textmesh} is similar to DreamFusion, but uses an implicit surface representation (\ie, VolSDF~\cite{yariv2021volume}).\\
\noindent\textbf{ProlificDreamer}~\cite{wang2023prolificdreamer} proposes to model the distribution of the possible 3D content given a text prompt, which differentiates it from others, where only a point estimate is obtained.\\
\noindent\textbf{MVDream}~\cite{shi2023mvdream} is detailed in section~\ref{sect-mvdream}.\\
\noindent\textbf{DreamGaussian}~\cite{tang2023dreamgaussian} proposes a two-stage approach, where the first stage is similar to DreamFusion, but uses Gaussian Splatting~\cite{kerbl20233d-gaussiansplatting} instead of NeRF, and its second stage conducts texture-only refinement on the textured mesh extracted from the first stage's representation.\\
\noindent\textbf{DreamGaussian+MVDream}. The authors of DreamGaussian recently open-sourced a version that integrated MVDream~\cite{shi2023mvdream}. To make our comparison more thorough and up-to-date, we also compare against this approach.\\
\noindent\textbf{GSGEN}~\cite{chen2023text-gsgen} is another recent Gaussian Splatting-based approach, which uses a pre-trained Point-E~\cite{nichol2022pointe} model to both initialize the Gaussians and regularize the training process.\\
\noindent\textbf{GaussianDreamer}~\cite{yi2023gaussiandreamer} is yet another Gaussian Splatting-based approach that uses Shap-E~\cite{jun2023shape} fine-tuned on the Objaverse dataset for initialization.

Since the original implementation of DreamFusion, Magic3D, TextMesh, and ProlificDreamer is not publicly available, we use their best open-source re-implementation in ThreeStudio~\cite{threestudio2023} (already used in \cite{shi2023mvdream,tang2023dreamgaussian,yi2023gaussiandreamer,chen2023text-gsgen}). We use the original implementation for the other methods.

\subsection{Used text prompts}
\label{sect-prompts}
We categorize our prompts into four categories: \textit{single object, compositional objects, single animal, and compositional animals}. Rationale behind categorizing objects and animals is that we observe the Janus problem more often in the animal category than objects. We adopt this categorization on the prompts utilized in DreamFusion~\cite{poole2023dreamfusion} since it offers the diverse and challenging prompts suitable for our evaluation protocol. In order to obtain our list of prompts, we randomly select 25 prompts from each category, resulting in total of 100 prompts. We include a complete list of the prompts in the supplementary material. 

\subsection{Quality metrics}
\label{sect-metrics}
To properly assess the quality of text-to-3D, we ask the following three questions:

\textit{First, how often does the Janus problem appear in the generated 3D content?} As mentioned in section~\ref{sect-intro} and section~\ref{sect-sds-issues}, the Janus problem is one of the major failure cases of SDS-based text-to-3D methods.
Even for simple prompts, very often, recent text-to-3D approaches suffer from the Janus problem.
We argue that, without measuring quantitatively the Janus problem, which is a catastrophic failure, and by only using visual comparisons of selected generated 3D models, it is not possible to demonstrate actual improvements.
Therefore, our first and foremost evaluation metric is the frequency of the Janus problem, that is, out of a relatively large number of trials (100 in our experiments), the percentage of trials that suffer from the Janus problem.
In our experiments, we observe that the severity of the Janus problem varies from method to method. Besides the obvious ones where the generated 3D content has multiple faces, there are also subtle cases where the generated 3D content has additional or missing body parts (\eg, a corgi with 3 ears). We count both cases as having the Janus problem, and show the evaluation results of many state-of-the-art methods along with ours in Table~\ref{tab-janus}. It is not hard to see from the results that all the baseline methods except those utilizing multiview diffusion models such as \texttt{DreamGaussian+MVDream} and \texttt{MVDream} encounters a very high frequency of the Janus problem -- oftentimes, more than half of the generated 3D content out of the 100 prompts that we have tried has this problem.
Our method has relatively low frequency of the Janus problem thanks to the MVDream component which itself encounters no Janus problems in our test. As discussed in section~\ref{sect-proposed-method}, our refinement stage utilizes supervision from Stable Diffusion for improved realism at the cost of reduced multiview consistency, and as a result, our full model has slightly higher Janus problem ratio (6\%) compared to our first stage model (1\%) -- still much lower than the majority of the baselines. The improved realism as well as text and 3D alignment can be seen in Table~\ref{tab-fidelity} and \ref{tab-alignment}. We found that PerpNeg~\cite{armandpour2023perpneg} that aims to resolve the Janus problem only reduces the frequency of the Janus problem from 69\% to 64\% compared to the base model \texttt{DreamFusion}, and still suffers from very severe Janus problems.

We want to point out that as a first step towards a proper evaluation protocol for text-to-3D, we are addressing the ``what to measure'' question.
Developing a reliable Janus problem detection algorithm for automatic evaluation (``how to measure'') is desirable, but is non-trivial and beyond the paper's scope. As such we leave it for future work. In this work, we manually inspect the Janus problem instead. Nevertheless, we discuss a few attempted automatic evaluation methods in the supplementary material.
\begin{table}[t]
  \centering
  \tiny
  \begin{tabular}{|l|c|c|c|}
    \hline
    Method                          &  Freq. of Janus (\%) $\downarrow$ \\
    \hline\hline
        \rowcolor{lightgray}\texttt{DreamFusion}~\cite{poole2023dreamfusion}             &   69  \\
    \rowcolor{lightgray}\texttt{DreamFusion+PerpNeg}~\cite{armandpour2023perpneg}    &   64   \\
    \rowcolor{lightgray}\texttt{Magic3D}~\cite{lin2023magic3d}                       &   69  \\
    \rowcolor{lightgray}\texttt{TextMesh}~\cite{tsalicoglou2023textmesh}             &   58  \\
    \rowcolor{lightgray}\texttt{ProlificDreamer}~\cite{wang2023prolificdreamer}      &   78  \\ 
    \rowcolor{lightgray}\texttt{MVDream}~\cite{shi2023mvdream}                       &   \textbf{0}   \\
    \rowcolor{SpringGreen}\texttt{DreamGaussian}~\cite{tang2023dreamgaussian}          &   87  \\
    \rowcolor{SpringGreen}\texttt{DreamGaussian+MVDream}~\cite{tang2023dreamgaussian}  &   8   \\
    \rowcolor{SpringGreen}\texttt{GSGEN}~\cite{chen2023text-gsgen}                     &   66  \\
    \rowcolor{SpringGreen}\texttt{GaussianDreamer}~\cite{yi2023gaussiandreamer}        &   46  \\
    \hline  
    \rowcolor{SpringGreen}\texttt{Ours -- 1st stage}      &   \underline{1}  \\     
    \rowcolor{SpringGreen}\texttt{Ours -- full model}     &   6   \\   
    \hline
  \end{tabular}
  \caption{\textit{Frequency of the Janus problem}. We run each method on 100 prompts to generate 3D content and manually count the frequency of the Janus problem. All the baseline methods except those utilizing multiview diffusion models such as \texttt{DreamGaussian+MVDream}, \texttt{MVDream}, and ours have very severe instances of the Janus problem. \textbf{Bold} represents the best result and \underline{underline} represents the second best. \colorbox{SpringGreen}{\phantom -} indicates Gaussian Splatting-based text-to-3D methods, and \colorbox{lightgray}{\phantom -} indicates methods that are not based on Gaussian Splatting.}
  \label{tab-janus}
  \vspace{-0.2in}
\end{table}

\begin{figure}[t]
\centering
\includegraphics[width=0.8\linewidth]{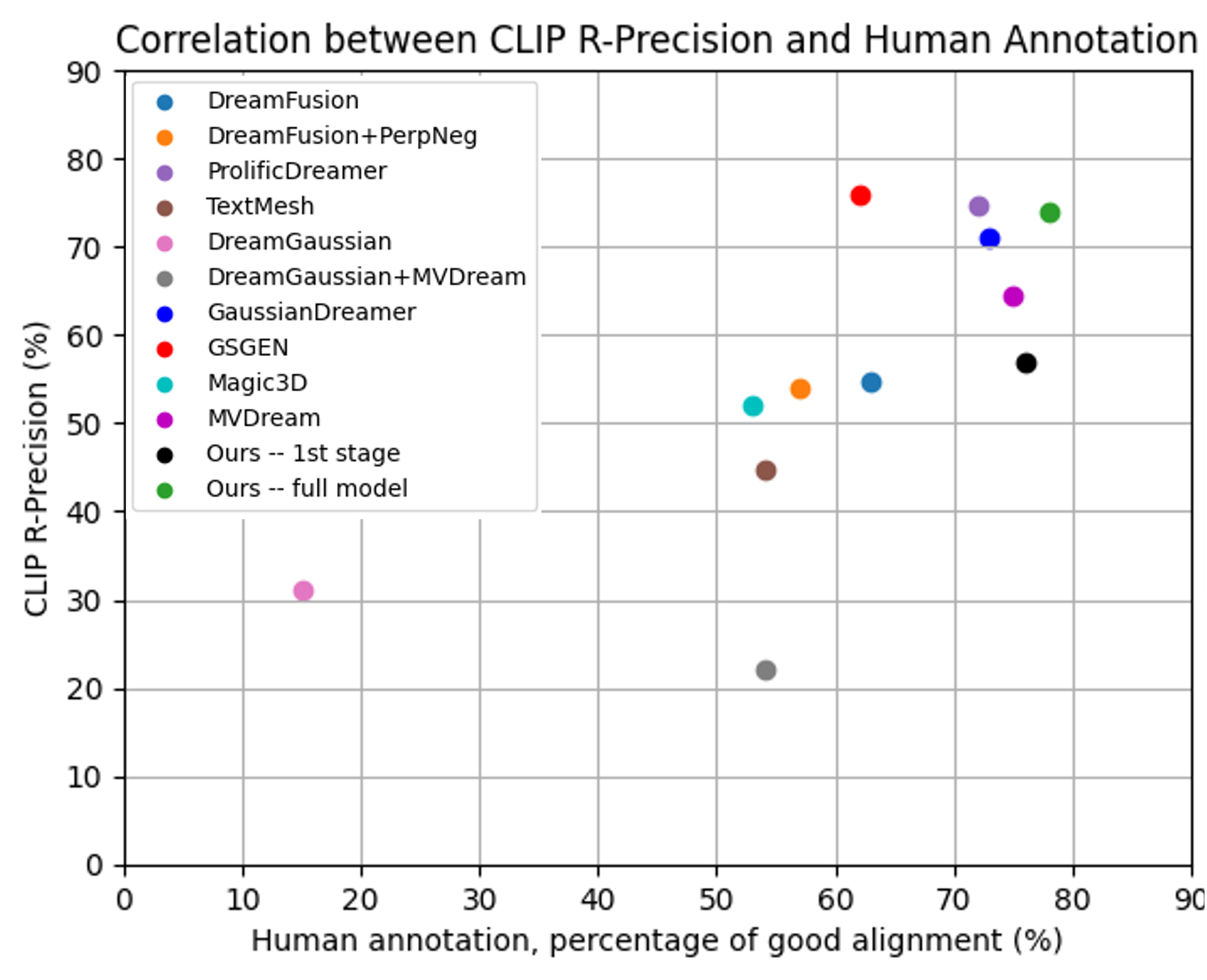}
\caption{\textit{Correlation between CLIP R-Precision and human annotation}. The abscissa shows the percentage of the generated 3D content that are annotated by human reviewers as aligning well with the given text prompts. The ordinate shows CLIP R-Precision. We can see from the plot that CLIP R-Precision correlates well with the human annotation, which supports our choice of the CLIP R-Precision as the algorithmic method for evaluating text-to-3D alignment.}
\label{fig-corr}
\vspace{-0.2in}
\end{figure}

\textit{Second, we ask if the generated 3D content aligns well with the given text prompt.} We consider two evaluation methods: human evaluation, and automatic evaluation based on CLIP R-Precision~\cite{jain2022zero,poole2023dreamfusion}. For the human evaluation, a human reviewer is asked to score each 3D model into 2 brackets: (i) good, meaning that the 3D model captures all the aspects of the prompt; (ii) poor, meaning that the human reviewer considers the prompt is poorly captured by the 3D model. For the automatic evaluation, we follow \cite{jain2022zero,poole2023dreamfusion}, and evaluate the CLIP R-Precision that measures the consistency of the 3D model's rendered images with respect to the input text prompt. The R-Precision is the accuracy with which CLIP~\cite{radford2021learning-clip} retrieves the correct prompt among a set of candidate text prompts given a rendering of the generated 3D model. Table~\ref{tab-alignment} shows both the human rating and CLIP R-Precision of various baselines and our method for text and 3D alignment. To validate that CLIP R-Precision is indeed a good metric for alignment, Figure~\ref{fig-corr} shows a correlation plot between the human rating and the CLIP R-Precision. Based on the results in Table~\ref{tab-alignment}, we found that \texttt{GSGEN}, \texttt{GaussianDreamer}, \texttt{MVDream}, and \texttt{ProlificDreamer} have relatively good alignment. We hypothesize that this is because both \texttt{GSGEN} and \texttt{GaussianDreamer} utilize 3D diffusion models as shape prior (Point-E~\cite{nichol2022pointe} and Shap-E~\cite{jun2023shape} respectively) in addition to the SDS framework leading to overall better geometry. And likewise, \texttt{MVDream} utilizes a multiview diffusion model which also results in better geometry. \texttt{ProlificDreamer} models the nuisance variabilities better by modeling a distribution instead of a point estimate of the 3D content. Ours performs on par with the top performing methods in both human and automatic evaluation.\\
\begin{table}[t]
  \centering
  \tiny
  \begin{tabular}{|l|c|c|}
    \hline
    Method &  Good alignment (\%) $\uparrow$ & CLIP R-Precision (\%) $\uparrow$ \\
    \hline\hline
    \rowcolor{lightgray}\texttt{DreamFusion}~\cite{poole2023dreamfusion}               &  63.00 & 54.64 \\
    \rowcolor{lightgray}\texttt{DreamFusion+PerpNeg}~\cite{armandpour2023perpneg}      &  56.00 & 54.00    \\
    \rowcolor{lightgray}\texttt{Magic3D}~\cite{lin2023magic3d}                         &  51.00 & 52.00 \\
    \rowcolor{lightgray}\texttt{TextMesh}~\cite{tsalicoglou2023textmesh}               &  55.00 & 44.75 \\
    \rowcolor{lightgray}\texttt{ProlificDreamer}~\cite{wang2023prolificdreamer}        &  72.00 & \underline{74.75} \\  
    \rowcolor{lightgray}\texttt{MVDream}~\cite{shi2023mvdream}                         &  73.00 & 64.37 \\
    \rowcolor{SpringGreen}\texttt{DreamGaussian}~\cite{tang2023dreamgaussian}            &  15.00 & 31.00 \\
    \rowcolor{SpringGreen}\texttt{DreamGaussian+MVDream}~\cite{tang2023dreamgaussian}    &  51.00 & 22.00 \\
    \rowcolor{SpringGreen}\texttt{GSGEN}~\cite{chen2023text-gsgen}                       &  64.00 & \textbf{76.00} \\
    \rowcolor{SpringGreen}\texttt{GaussianDreamer}~\cite{yi2023gaussiandreamer}          &  74.00 & 71.00 \\
    \hline
    \rowcolor{SpringGreen}\texttt{Ours -- 1st stage}                                     &  \underline{76.00} & 57.00 \\  
    \rowcolor{SpringGreen}\texttt{Ours -- full model}                                    &  \textbf{78.00} & 74.00 \\  
    \hline
  \end{tabular}
  \caption{\textit{Evaluation of text and 3D alignment}. We use both human rating and CLIP R-Precision. Under ``Good alignment'', for each method, we show the percentage of the generated 3D models that human reviewers annotate as aligning well with the text prompts. In Figure~\ref{fig-corr}, we show that the human annotation and the CLIP R-Precision do correlate with each other. For both human evaluation and CLIP R-Precision, ours is the most competitive among the state-of-the-art methods compared. \textbf{Bold}
represents the best result and \underline{underline} represents the second best.}
  \label{tab-alignment}
  \vspace{-0.1in}
\end{table}

\textit{The third question we ask is if the generated 3D content is realistic.} To this end, we compute the FID score~\cite{heusel2017gans}, which measures the distance between two collections of images. In our case, we form the first image collection by rendering images from the generated 3D models at evenly spaced azimuth angles, and the second image collection, which is considered as the (pseudo) ground truth, consists of samples generated by Stable Diffusion using the same set of text prompts that are used to generate the 3D models. Though Stable Diffusion is far from being perfect, we consider its generation quality superior to the best-in-class text-to-3D models, and as such, we believe it is reasonable to use samples drawn from Stable Diffusion as (pseudo) ground truth to measure the realism of text-to-3D models. In addition to the FID score, we also use the inception score~\cite{salimans2016improved} (IS) -- which is reference free -- as another realism metric. Table~\ref{tab-fidelity} shows the evaluation of FID and IS complemented by a qualitative evaluation in Figure~\ref{fig-qualitative-comparison}. As observed in the evaluation results, \texttt{MVDream}, \texttt{ProlificDreamer}, and ours have the best fidelity among the methods compared. Our first stage model that simply integrates MVDream and Gaussian Splatting performs reasonably well (on par with \texttt{GaussianDreamer} which is another Gaussian Splatting-based method), but falls behind MVDream showing that a simple integration is insufficient to achieve realistic generation. The strong performance of our full model shows the necessity of the refinement stage.
\begin{table}[t]
  \centering
  \tiny
  \begin{tabular}{|l|c|c|}
    \hline
    Method &  FID $\downarrow$ & IS $\uparrow$\\
    \hline\hline
    \rowcolor{lightgray}\texttt{DreamFusion}~\cite{poole2023dreamfusion}             & 116.60             & 15.34 \\
    \rowcolor{lightgray}\texttt{DreamFusion+PerpNeg}~\cite{armandpour2023perpneg}    & 128.06             & 13.02     \\
    \rowcolor{lightgray}\texttt{Magic3D}~\cite{lin2023magic3d}                       & 130.16             & 12.41 \\
    \rowcolor{lightgray}\texttt{TextMesh}~\cite{tsalicoglou2023textmesh}             & 129.87             & 11.05 \\
    \rowcolor{lightgray}\texttt{ProlificDreamer}~\cite{wang2023prolificdreamer}      & 83.89              & \textbf{22.33}  \\
    \rowcolor{lightgray}\texttt{MVDream}~\cite{shi2023mvdream}                       & \underline{80.99}  & 18.32 \\
    \rowcolor{SpringGreen}\texttt{DreamGaussian}~\cite{tang2023dreamgaussian}          & 191.99             & 4.32  \\
    \rowcolor{SpringGreen}\texttt{DreamGaussian+MVDream}~\cite{tang2023dreamgaussian}  & 158.94             & 5.82  \\
    \rowcolor{SpringGreen}\texttt{GSGEN}~\cite{chen2023text-gsgen}                     & 104.45             & 13.66 \\
    \rowcolor{SpringGreen}\texttt{GaussianDreamer}~\cite{yi2023gaussiandreamer}        & 93.25              & 17.41 \\
    \hline
    \rowcolor{SpringGreen}\texttt{Ours -- 1st stage}                                   & 94.16              & 12.79 \\
    \rowcolor{SpringGreen}\texttt{Ours -- full model}                                  & \textbf{79.45}     & \underline{18.44} \\
    \hline
  \end{tabular}
  \caption{\textit{Evaluation of image realism via FID and IS}. Our method yields competitive FID and IS scores among all the compared methods. \textbf{Bold}
represents the best result and \underline{underline} represents the second best.}
  \label{tab-fidelity}
  \vspace{-0.2in}
\end{table}

\subsection{Training efficiency metric}
\label{sect-efficiency}
\begin{figure*}[h]
     \centering
    \includegraphics[width=0.80\linewidth]{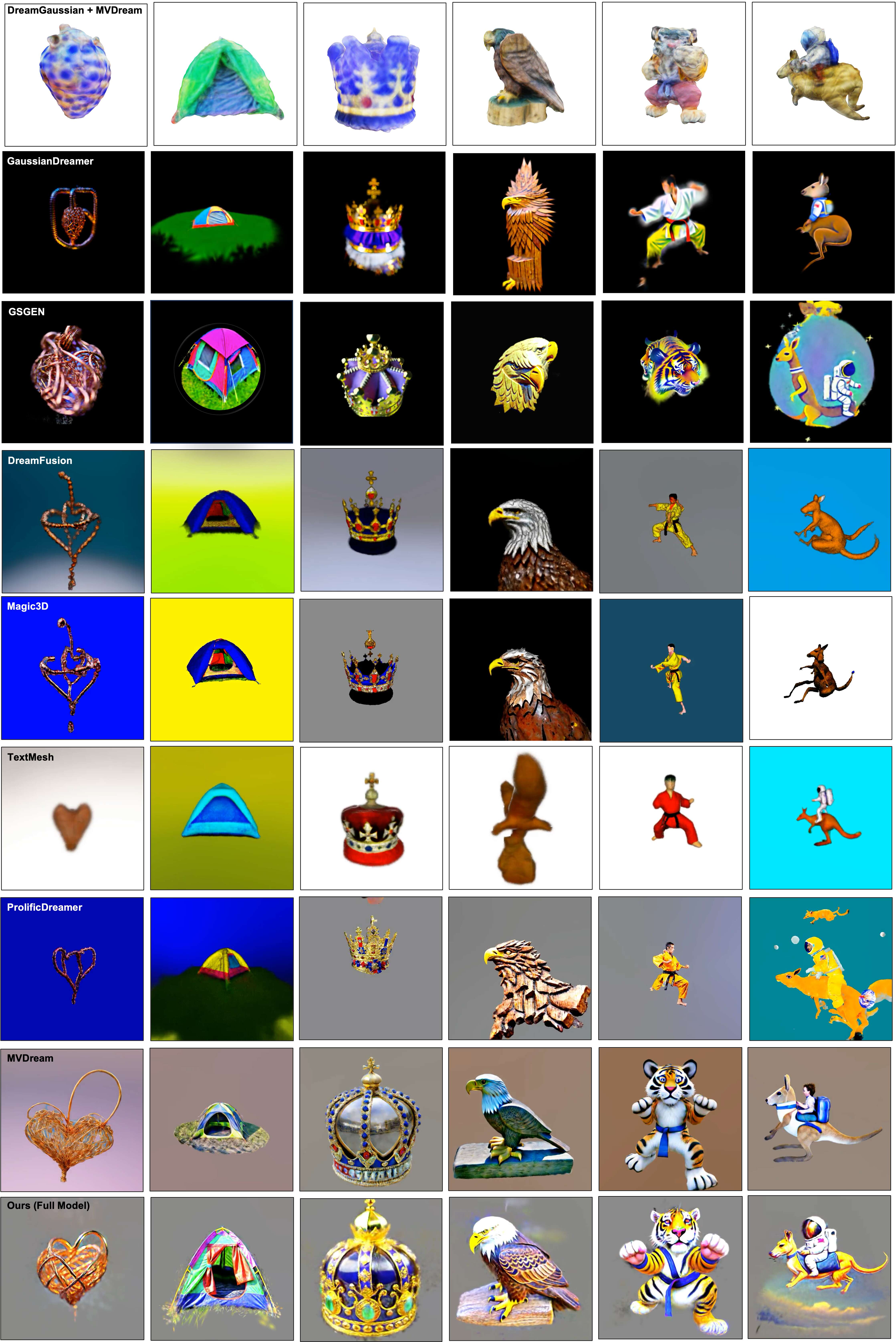}
    \caption{\footnotesize\textit{Qualitative comparison.} Each row shows 3D models generated by one method using different prompts.
    Prompts used (left to right): ``a DSLR photo of A very beautiful tiny human heart organic sculpture made of copper wire and threaded pipes, very intricate, curved, Studio lighting, high resolution'', 
    ``a zoomed out DSLR photo of a colorful camping tent in a patch of grass'',,
    ``a DSLR photo of the Imperial State Crown of England'',
    ``a bald eagle carved out of wood'',
    ``a tiger karate master'',
    ``an astronaut riding a kangaroo''.
    Due to the space limit, we do not include results of \texttt{DreamFusion+PerpNeg} (similar to \texttt{DreamFusion}) and \texttt{DreamGaussian} (very poor compared to its updated version \texttt{DreamGaussian+MVDream}). More visual results can be found in the supplementary material.}
    \label{fig-qualitative-comparison}
\end{figure*}

For text-to-3D methods to be useful in real-world applications, the time they need to generate a 3D model must be reasonable. To this end, we also introduce an efficiency metric for the training process. Specifically, we measure the GPU-hours needed by each method to generate one 3D model. For a fair comparison, all the GPU-hours are measured on Nvidia A100 GPUs, and we use the hyperparameters recommended by the authors of each method.
Table~\ref{tab-efficiency} shows the results. We observe that \texttt{DreamGaussian} and its variant \texttt{DreamGaussian+MVDream} are the most efficient among the methods compared. Yet, we found that their generation quality is often the poorest as measured by the quality metrics (see Table~\ref{tab-janus}, Table~\ref{tab-alignment}, and Figure~\ref{fig-qualitative-comparison}). This is probably due to DreamGaussian's extreme optimization for efficiency over quality. Among the rest of the methods, our proposed method is as efficient as \texttt{GaussianDreamer} -- another recent Gaussian Splatting-based method, and at least twice more efficient than the non-Gaussian Splatting based methods.
\begin{table}[hbt]
  \centering
  \tiny
  \begin{tabular}{|l|c|}
    \hline
    Method &  GPU-hours $\downarrow$\\
    \hline\hline
    \rowcolor{lightgray}\texttt{DreamFusion}~\cite{poole2023dreamfusion}             &  1.5   \\
    \rowcolor{lightgray}\texttt{DreamFusion+PerpNeg}~\cite{armandpour2023perpneg}    &  1.5   \\
    \rowcolor{lightgray}\texttt{Magic3D}~\cite{lin2023magic3d}                       &  2.4   \\
    \rowcolor{lightgray}\texttt{TextMesh}~\cite{tsalicoglou2023textmesh}             &  1.26  \\
    \rowcolor{lightgray}\texttt{ProlificDreamer}~\cite{wang2023prolificdreamer}      &  15.5  \\
    \rowcolor{lightgray}\texttt{MVDream}~\cite{shi2023mvdream}                       &  1.83  \\
    \rowcolor{SpringGreen}\texttt{DreamGaussian}~\cite{tang2023dreamgaussian}          &  \textbf{0.08}  \\
    \rowcolor{SpringGreen}\texttt{DreamGaussian+MVDream}~\cite{tang2023dreamgaussian}  &  \textbf{0.08}  \\
    \rowcolor{SpringGreen}\texttt{GSGEN}~\cite{chen2023text-gsgen}                     &  1.67  \\
    \rowcolor{SpringGreen}\texttt{GaussianDreamer}~\cite{yi2023gaussiandreamer}        &  0.42  \\
    \hline    
    \rowcolor{SpringGreen}\texttt{Ours -- 1st stage}                                   &  \underline{0.34}  \\
    \rowcolor{SpringGreen}\texttt{Ours -- full model}                                  &  0.50  \\
    \hline
  \end{tabular}
  \caption{\textit{GPU-hours needed to generate one 3D model given a prompt}. \texttt{DreamGaussian} and its multiview variant \texttt{DreamGaussian+MVDream} are the most efficient among all the methods compared. However, they fall behind on other metrics (see Table~\ref{tab-alignment},~\ref{tab-fidelity}, and Figure~\ref{fig-qualitative-comparison}). Among the rest of the methods, ours is as efficient as \texttt{GaussianDreamer}, and at least twice more efficient than the others. \textbf{Bold}
represents the best result and \underline{underline} represents the second best.}
  \label{tab-efficiency}
  \vspace{-0.1in}
\end{table}

\section{Related work}
\label{sect-related}
Score Distillation Sampling (SDS) is proposed by Boole~\etal in DreamFusion~\cite{poole2023dreamfusion} and followed by many~\cite{wang2023sjc,lin2023magic3d,tsalicoglou2023textmesh,metzer2023latent,chen2023fantasia3d,wang2023prolificdreamer, shi2023mvdream,tang2023dreamgaussian,chen2023text-gsgen,yi2023gaussiandreamer,li2023sweetdreamer,sun2023dreamcraft3d}. The key idea is to supervise the training of a NeRF~\cite{mildenhall2021nerf,barron2021mip} using the supervision from a pre-trained and frozen large text-to-image model~\cite{rombach2022high-ldm,saharia2022photorealistic_imagen}. Several techniques have been proposed to improve the SDS framework including better viewpoint conditioning~\cite{armandpour2023perpneg}, better timestep scheduling~\cite{huang2023dreamtime}, variational score distillation~\cite{wang2023prolificdreamer}, accelerated NeRF representation~\cite{muller2022instant}, surface representation~\cite{yariv2021volume,wang2021neus,shen2021deep-dmtet}, improved efficiency using Gaussian Splatting~\cite{kerbl20233d-gaussiansplatting}, and improved fidelity~\cite{zhu2023hifa}. 
In contrast to the SDS framework, another paradigm in 
text-to-3D is to directly learn the mapping from texts to 3D using diffusion models~\cite{nichol2022pointe,jun2023shape} on large-scale 3D dataset such as Objaverse~\cite{deitke2023objaverse}.
Another line of work that is related to text-to-3D is single image to 3D -- notable ones includes \cite{tang2023make,melas2023realfusion,qian2023magic123,liu2023one2345,lin2023consistent123,liu2023syncdreamer,long2023wonder3d}.
Regarding evaluation, $\mathrm{T}^3$Bench~\cite{he2023t3bench} -- which is concurrent to our work -- proposes a fairly sophisticated protocol to evaluate the quality and alignment of text-to-3D by utilizing LLMs~\cite{li2022blip,openai2023gpt}. Though $\mathrm{T}^3$Bench mentions that its approach could be used to detect the Janus problem, it never shows how effective it is, and does not single out and characterize the Janus problem, which, unfortunately, seems to be one of the major failure cases of text-to-3D.

\section{Conclusion}
\label{sect-conclusion}
We proposed an evaluation protocol to examine three key aspects of text-to-3D models: the Janus problem, the text and 3D alignment, and the realism of the generated 3D content. By using this protocol, we evaluated several state-of-the-art methods, and were able to characterize the limitations of these methods. Through these findings, we proposed a new text-to-3D model that is efficient and performs well on all the quality metrics, thus setting a strong baseline for future text-to-3D work. Future directions include further improving the efficiency of text-to-3D, leveraging both real-world and synthetic data to further improve the diversity, alignment, and realism of 3D content generation.

\clearpage
\setcounter{page}{1}
\maketitlesupplementary


\section{Additional ablation results on refinement stage}
\label{sec:ablation-refinement}

\begin{figure}[t]
\centering
\includegraphics[width=1.0\linewidth]{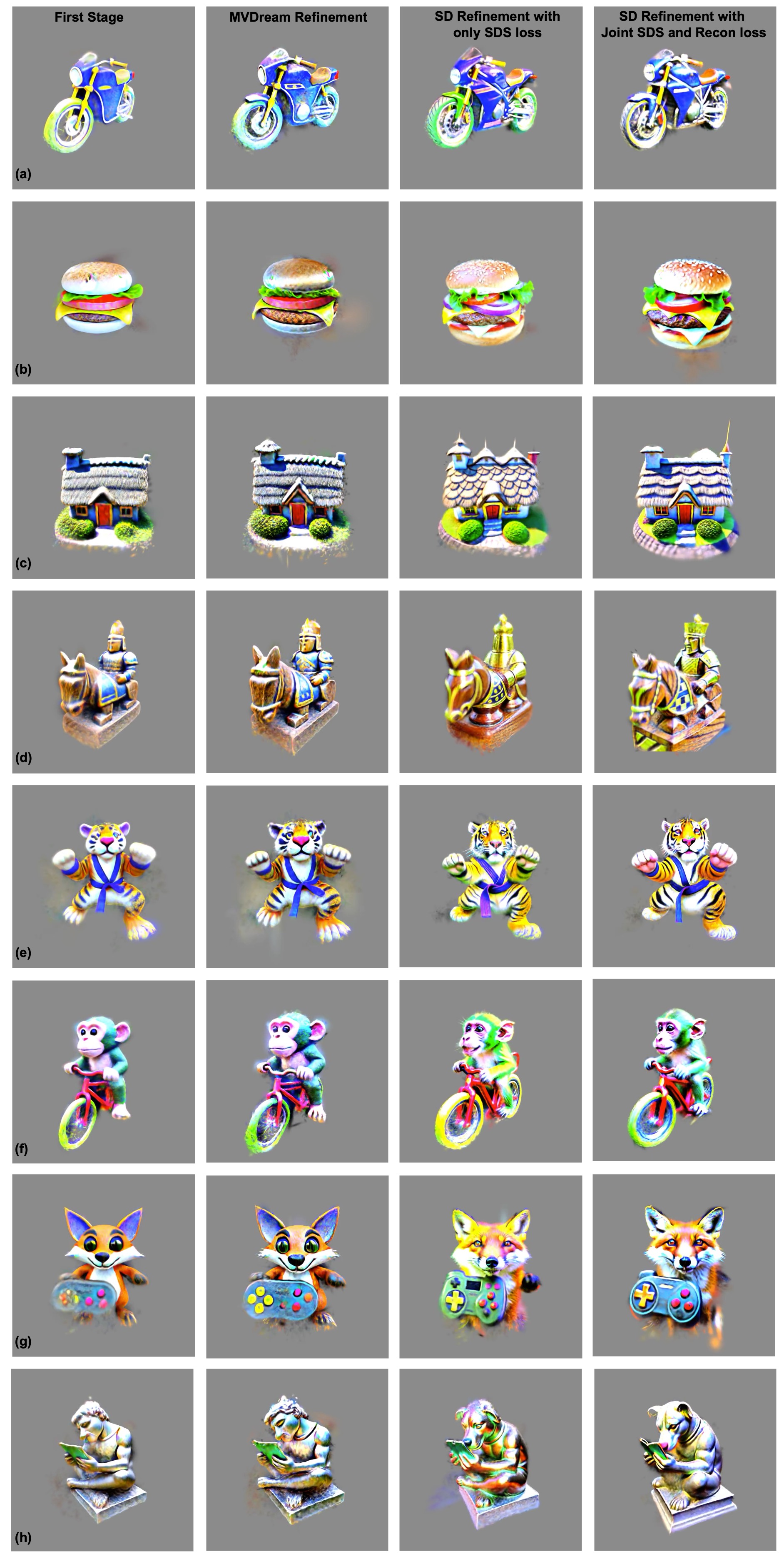}
\vspace{-0.2in}
\caption{\footnotesize\textit{Additional ablation of the refinement stage.} Each row shows 3D models generated by different prompts.
Prompts used : (a) ``a blue motorcycle", (b) ``a delicious hamburger", (c)  ``a zoomed out DSLR photo of a 3d model of an adorable cottage with a thatched roof", (d) ``a zoomed out DSLR photo of a beautifully carved wooden knight chess piece", (e) ``a tiger karate master", (f) ``a zoomed out DSLR photo of a monkey riding a bike", (g) ``a DSLR photo of a fox holding a videogame controller", (h) ``Michelangelo style statue of dog reading news on a cellphone". Each column demostrates the results from various stages (Left - Right): Our first stage, MVDream only refinement, Stable Diffusion only refinement, refinement with the joint loss using both Stable Diffusion and MVDream (proposed in the main text).}
\label{fig-supp_refinement_ablation}
\vspace{-0.1in}
\end{figure}

In this section, we present additional ablation studies of the refinement stage. In contrast to Figure~\ref{fig-secondstage} of the main text where we compare our first stage and full model that utilizes both Stable Diffusion and MVDream for refinement (in the form of SDS loss through Eq.~\eqref{eq-sds-loss} and Eq.~\eqref{eq-sds-mvdream}), here we showcase two alternative design choices in the refinement stage: (1) MVDream only, and (2) Stable Diffusion only.
Figure~\ref{fig-supp_refinement_ablation} shows the comparison. Each row shows (renderings of) the 3D models generated by different prompts. In each column, from left to right, we show the first stage results, refinement using only MVDream, refinement using only Stable Diffusion, and refinement using both MVDream and Stable Diffusion which is what we propose in the main text. For MVDream only refinement (second column), after the first stage, we train the Gaussian Splatting model with MVDream for additional 1000 iterations. We observe that, even with additional training, MVDream only refinement cannot improve the photorealism and diversity as compared to the joint refinement. In the third column (Stable Diffusion only refinement), thanks to Stable Diffusion's generalization and photorealism capability, we observe the improved quality and text to 3D alignment -- especially in burger (b) and Michelangelo style statue (h), thereby filling the domain gap. However, we also observe that the generated models tend to diverge slightly from the results of the first stage and acquire different colors and textures. Our joint loss (last column) helps in preserving the colors and textures while improving the fidelity and photorealism.

\section{Trade-off between Janus and Quality}
\label{sec:janus-and-fidelity}

\begin{figure*}[h]
     \centering
    \includegraphics[width=0.9\linewidth]{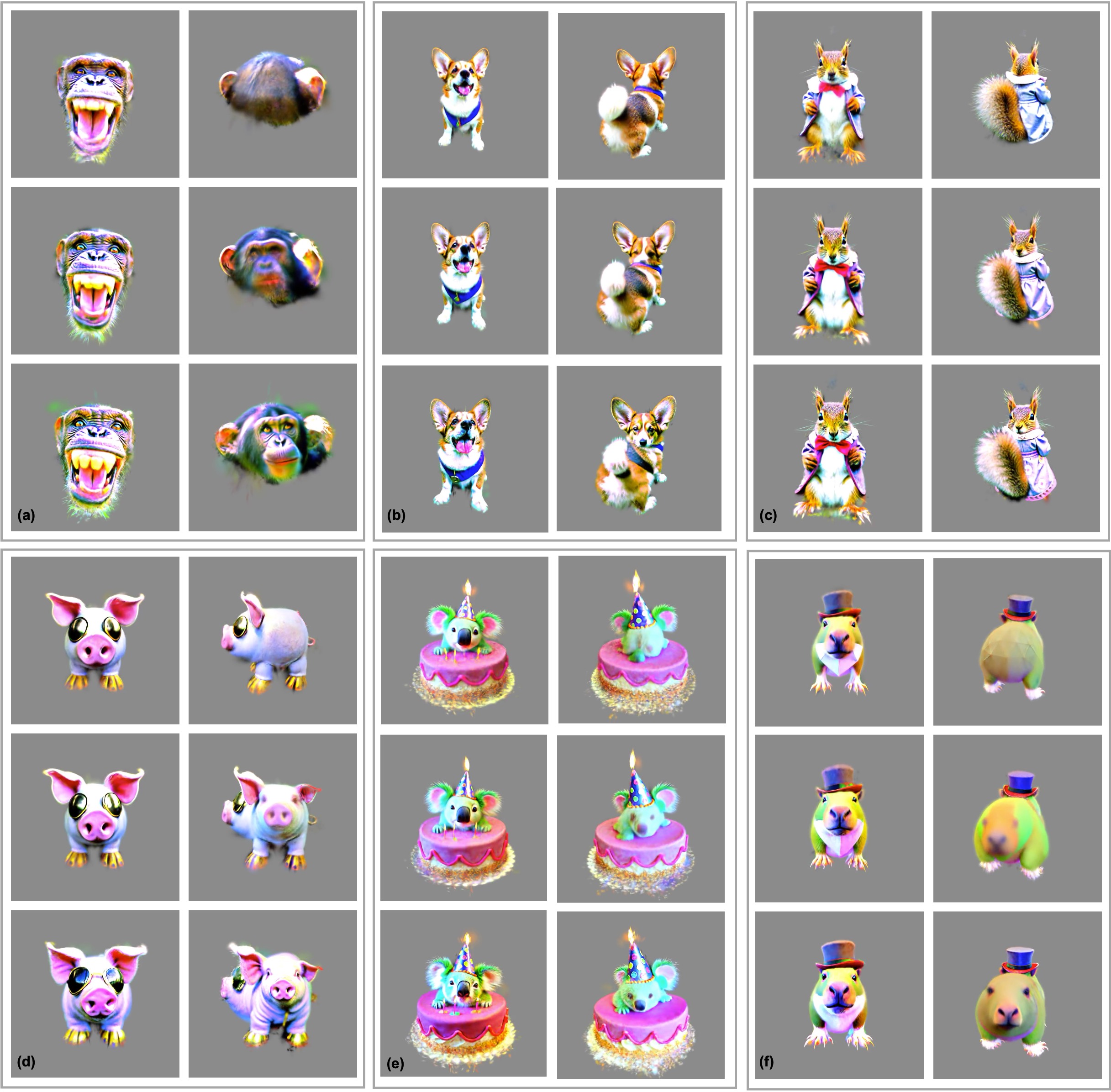}
    \caption{\footnotesize\textit{Ablation on steps of the refinement stage for trade-off between Janus and Quality.} 
    Prompts used : (a) ``a chimpanzee with a big grin",
    (b) ``a DSLR photo of a corgi puppy",
    (c) ``a zoomed out DSLR photo of a squirrel dressed up like a Victorian woman",
    (d) ``a pig wearing a backpack",
    (e) ``a DSLR photo of a koala wearing a party hat and blowing out birthday candles on a cake",
    (f) ``a capybara wearing a top hat, low poly". In each prompt (a)-(f), the multi-view images correspond to refinement stage with 300 iterations (first row), 500 iterations (second row), and 1000 iterations (third row). We observe the quality improves (addition of finer details, improvement in sharpness) as the Janus problem emerges.}
    \label{fig-supp_janus_fidelity_ablation}
\end{figure*}

The one potential risk of using Stable Diffusion together with MVDream in our refinement stage is the emergence of the Janus problem.
Though Stable Diffusion generates more realistic images compared to MVDream which is trained mostly on synthetic data~\footnote{MVDream is trained on a mixture of LAION real-world images (30\%) and multiview images rendered using the Objaverse 3D dataset (70\%).}, its generations are {\em not} multiview consistent compared to MVDream.
Hence, our joint refinement with Stable Diffusion and MVDream has a trade-off between the occurrence of the Janus problem and the quality of the generated 3D models.
In this section, we study this trade-off. We increase the number of refinement steps in Figure~\ref{fig-supp_janus_fidelity_ablation}. As we increase the steps from 300 to 1000, we encounter incremental addition of the Janus issue (from no-Janus to Janus). With the help of visual inspection, on the other hand, we observe that with more refinement steps, the visual quality of the generated 3D models has been improved: at 300 steps, the quality of the generated 3D model is slightly worse (blurriness, lack of fine details) compared to the generated 3D model at 1000 steps. 
These observations confirm the trade-off, and explain the increased frequency of the Janus problem of our full model (compared to our first stage only model) in Table~\ref{tab-janus} of the main text.

\section{Automatic evaluation of Janus problem}
\label{sec:automatic-evaluation}

In the main text, we opted for manual inspection of the Janus problem due to the lack of good automatic evaluation method. In this section, we discuss a few approaches that we have tried for automatic detection of the Janus problem.

The Janus problem happens when the same view (often the face of an animal) is incorrectly repeated multiple times across different viewpoints. That said, the key to detect the Janus problem is to detect the repeated faces -- at least for the animal cases. For instance, assuming the first frame in the rendered video corresponds to the front view of an animal where the face is visible, if we can detect other views that also contain the face of the animal, we will be able to detect the Janus problem. But how do we detect the face?

A naive solution will be to train a face detector for all possible object categories which is apparently {\em not scalable}. An alternative solution is to conduct one-shot or few-shot detection, that is, using a few reference images (for example, the first few frames in the rendered video) as the ground truth, and detect views that are similar to the ground truth in the rendered video.

To this end, we have developed a simple detector to determine if a given frame is similar to the first frame of the video where the video is rendered by varying the azimuth angle from 0 to 360 degrees while fixing the elevation angle and distance of the camera to the object center. We first extract visual features (we have tried both DINO-v2 and CLIP features) from each of the frame in the rendered video. We then compute the distance of each frame to the reference frame (the first frame) in the feature space resulting in a list of distances $[dist_1, dist_2, \cdots, dist_K]$ where $K$ is the number of rendered frames. $dist_1$ is zero as it is the distance of the first frame to itself. $dist_2$ and $dist_K$ should be relatively small as they are rendered at viewpoints that are close to the first frame's viewpoint. We use $\tau = \rho \cdot (dist_2 + dist_K)$ as a threshold to determine if a given frame indexed by $k$ is similar to the reference frame where $\rho$ is a tuning parameter by varying which we can trade off the precision and recall of the detector. If the distance $dist_k$ of the $k$-th frame to the reference frame is smaller than the threshold, we consider it as similar to the reference frame. By thresholding the distance to the reference frame, we obtain a list of zeros and ones where a one at the $k$-th position of the list indicates that the $k$-th frame is similar to the reference frame, and zero otherwise. Ideally, if there is no Janus problem, we will only have consecutive ones at the head and tail of the list, as the frames at the beginning and end of the video are rendered at viewpoints close to the reference frame's viewpoint, and thus these frames should be similar to the reference frame. We can also imagine that if a face appears on the back of the generated 3D model, we will be able to detect a sub-sequence of ones in the middle of the list.

We have implemented the above mentioned detector and tested it on the rendered videos of the generated 3D models. Since we have manually annotated each video as having Janus problem or not, we can use these annotation to evaluate our detector. As mentioned above, by varying the tuning parameters, we can trade off the precision and recall of our detector. However, the best accuracy we could achieve is around 50\% -- not reliable enough to be used in automatic detection of the Janus problem.

We hypothesize that this is probably due to the fact that off-the-shelf feature extractors (DINO-v2, CLIP) do not capture part-level semantic information well. For instance, intuitively, one would be able to tell if an image contains a face or not by checking semantically meaningful facial landmarks such as eyes, ears, nose, and mouth. Yet, it's unclear if DINO-v2 or CLIP are able to capture meaningful features at that level of granularity. That said, another possible way to detect the Janus problem is to use image captioning models to describe what's visible in the image (it may be able to find the aforementioned facial landmarks), and then use the captioning results as features~\footnote{For example, one could use one-hot encoding of the facial landmarks as the feature.} to construct the Janus problem detector. In our preliminary experiments, we didn't succeed in this direction either as we were not able to get reliable part-level descriptions. We leave more in-depth investigation and development of automatic Janus problem detection as future work.

\section{Implementation details}
\label{sec:add-implementation-details}

In this section, we present the additional implementation details. All our experiments are carried out on A100 GPU with batch size of 12. We render our generated model at 256$\times$256 resolution. During the training, we set the camera distance range as [0.8, 1.2], camera FOV range as [15, 60] degrees, and elevation range as [-20, 60] degrees. During the evaluation, we set our camera distance as 3.0 and FOV as 40 degrees. For Gaussian Splatting, we utilize the learning rate for position as 0.0001, feature as 0.01, opacity as 0.003, scaling as 0.003, and rotation as 0.003. We set the gradient threshold for densification at 0.02 and opacity pruning threshold as 0.05. We adopt the compactness densification technique proposed in \cite{chen2023text-gsgen} for optimizing the number of gaussians and improving the rendering quality. 

In the first stage, we set MVDream guidance maximum and minimum time step percent as 0.98 and 0.02 respectively. And, in the refinement stage, we set Stable Diffusion guidance maximum and minimum time step percent as 0.5 and 0.2 respectively. Inspired by PrepNeg \cite{armandpour2023perpneg}, we utilize the negative prompting technique for improving the fidelity. The use of negative prompts (such as ``unrealistic, blurry, low quality, out of focus, low contrast, low-resolution") in the SDS loss can be seen as a dual-objective optimization that not only encourages the 3D renderings toward the positive prompt but also penalizes when the model generates undesired output. 

\section{List of text prompts}
\label{sec:prompts-list}
\small
\begin{enumerate}
\item     ``a DSLR photo of a car made out of cheese"
\item     ``a blue motorcycle"
\item     ``a completely destroyed car"
\item     ``a DSLR photo of A car made out of sushi"
\item     ``a DSLR photo of a green monster truck"
\item     ``a DSLR photo of a delicious croissant"
\item     ``a delicious hamburger"
\item     ``a zoomed out DSLR photo of an expensive office chair"
\item     ``a zoomed out DSLR photo of a 3d model of an adorable cottage with a thatched roof"
\item     ``a yellow schoolbus"
\item     ``a typewriter"
\item     ``a ripe strawberry"
\item     ``a flower made out of metal"
\item     ``a DSLR photo of the Imperial State Crown of England"
\item     ``a DSLR photo of a toy robot"
\item     ``a DSLR photo of a stack of pancakes covered in maple syrup"
\item     ``a DSLR photo of a Space Shuttle"
\item     ``a DSLR photo of an old vintage car"
\item     ``a DSLR photo of an ice cream sundae"
\item     ``a DSLR photo of a football helmet"
\item     ``a DSLR photo of a bulldozer"
\item     ``a DSLR photo of a baby grand piano viewed from far away"
\item     ``an airplane made out of wood"
\item     ``a zoomed out DSLR photo of a blue tulip"
\item     ``a zoomed out DSLR photo of Sydney opera house, aerial view"
\item     ``a beautiful dress made out of fruit, on a mannequin. Studio lighting, high quality, high resolution"
\item     ``a beautiful dress made out of garbage bags, on a mannequin. Studio lighting, high quality, high resolution"
\item     ``a bunch of colorful marbles spilling out of a red velvet bag"
\item     ``a DSLR photo of a bagel filled with cream cheese and lox"
\item     ``a DSLR photo of a beautiful violin sitting flat on a table"
\item     ``a DSLR photo of a candelabra with many candles on a red velvet tablecloth"
\item     ``a DSLR photo of a Christmas tree with donuts as decorations"
\item     ``a DSLR photo of a cup full of pens and pencils"
\item     ``a DSLR photo of a delicious chocolate brownie dessert with ice cream on the side"
\item     ``a DSLR photo of a humanoid robot using a laptop"
\item     ``fries and a hamburger"
\item     ``a zoomed out DSLR photo of a beautifully carved wooden knight chess piece"
\item     ``an orchid flower planted in a clay pot"
\item     ``a nest with a few white eggs and one golden egg"
\item     ``a DSLR photo of Two locomotives playing tug of war"
\item     ``a DSLR photo of a wooden desk and chair from an elementary school"
\item     ``a DSLR photo of a very cool and trendy pair of sneakers, studio lighting"
\item     ``a DSLR photo of A very beautiful tiny human heart organic sculpture made of copper wire and threaded pipes, very intricate, curved, Studio lighting, high resolution"
\item     ``a DSLR photo of a quill and ink sitting on a desk"
\item     ``a DSLR photo of a pair of headphones sitting on a desk"
\item     ``a plate of delicious tacos"
\item     ``a DSLR photo of a barbecue grill cooking sausages and burger patties"
\item     ``a zoomed out DSLR photo of a colorful camping tent in a patch of grass"
\item     ``a zoomed out DSLR photo of a few pool balls sitting on a pool table"
\item     ``a zoomed out DSLR photo of a bowl of cereal and milk with a spoon in it"
\item     ``a zoomed out DSLR photo of a blue lobster"
\item     ``a zoomed out DSLR photo of a baby dragon"
\item     ``a zoomed out DSLR photo of a kingfisher bird"
\item     ``a zoomed out DSLR photo of a corgi wearing a top hat"
\item     ``a ceramic lion"
\item     ``a bald eagle carved out of wood"
\item     ``a beagle in a detective's outfit"
\item     ``a beautiful rainbow fish"
\item     ``a chimpanzee with a big grin"
\item     ``a dragon-cat hybrid"
\item     ``a DSLR photo of a bald eagle"
\item     ``a DSLR photo of a bear dressed as a lumberjack"
\item     ``a DSLR photo of a corgi puppy"
\item     ``a DSLR photo of a human skull"
\item     ``a zoomed out DSLR photo of a shiny beetle"
\item     ``a zoomed out DSLR photo of a squirrel dressed up like a Victorian woman"
\item     ``a dachsund dressed up in a hotdog costume"
\item     ``a tiger karate master"
\item     ``a red panda"
\item     ``a plush toy of a corgi nurse"
\item     ``a lionfish"
\item     ``a highland cow"
\item     ``a capybara wearing a top hat, low poly"
\item     ``a DSLR photo of a pomeranian dog"
\item     ``a DSLR photo of a pirate collie dog, high resolution"
\item     ``a wide angle zoomed out DSLR photo of A red dragon dressed in a tuxedo and playing chess. The chess pieces are fashioned after robots"
\item     ``a pig wearing a backpack"
\item     ``a monkey-rabbit hybrid"
\item     ``Michelangelo style statue of dog reading news on a cellphone"
\item     ``a zoomed out DSLR photo of a tiger eating an ice cream cone"
\item     ``an astronaut riding a kangaroo"
\item     ``a DSLR photo of a koala wearing a party hat and blowing out birthday candles on a cake"
\item     ``a crocodile playing a drum set"
\item     ``a blue poison-dart frog sitting on a water lily"
\item     ``a bumblebee sitting on a pink flower"
\item     ``a DSLR photo of a corgi wearing a beret and holding a baguette, standing up on two hind legs"
\item     ``a DSLR photo of a ghost eating a hamburger"
\item     ``a zoomed out DSLR photo of a monkey riding a bike"
\item     ``a zoomed out DSLR photo of a raccoon astronaut holding his helmet"
\item     ``a zoomed out DSLR photo of a hippo biting through a watermelon"
\item     ``a zoomed out DSLR photo of a dachsund riding a unicycle"
\item     ``a zoomed out DSLR photo of a chimpanzee wearing headphones"
\item     ``a zoomed out DSLR photo of a beagle eating a donut"
\item     ``a ceramic upside down yellow octopus holding a blue green ceramic cup"
\item     ``a zoomed out DSLR photo of a pig playing the saxophone"
\item     ``a DSLR photo of a fox holding a videogame controller"
\item     ``a tiger playing the violin"
\item     ``a zoomed out DSLR photo of a chimpanzee holding a cup of hot coffee"
\item     ``a zoomed out DSLR photo of an adorable kitten lying next to a flower"
\item     ``a zoomed out DSLR photo of a rabbit digging a hole with a shovel"

\end{enumerate}
\section{Additional qualitative results}
\label{sec:add-qualitative-results}

\begin{figure*}[h]
     \centering
    \includegraphics[width=0.93\linewidth]{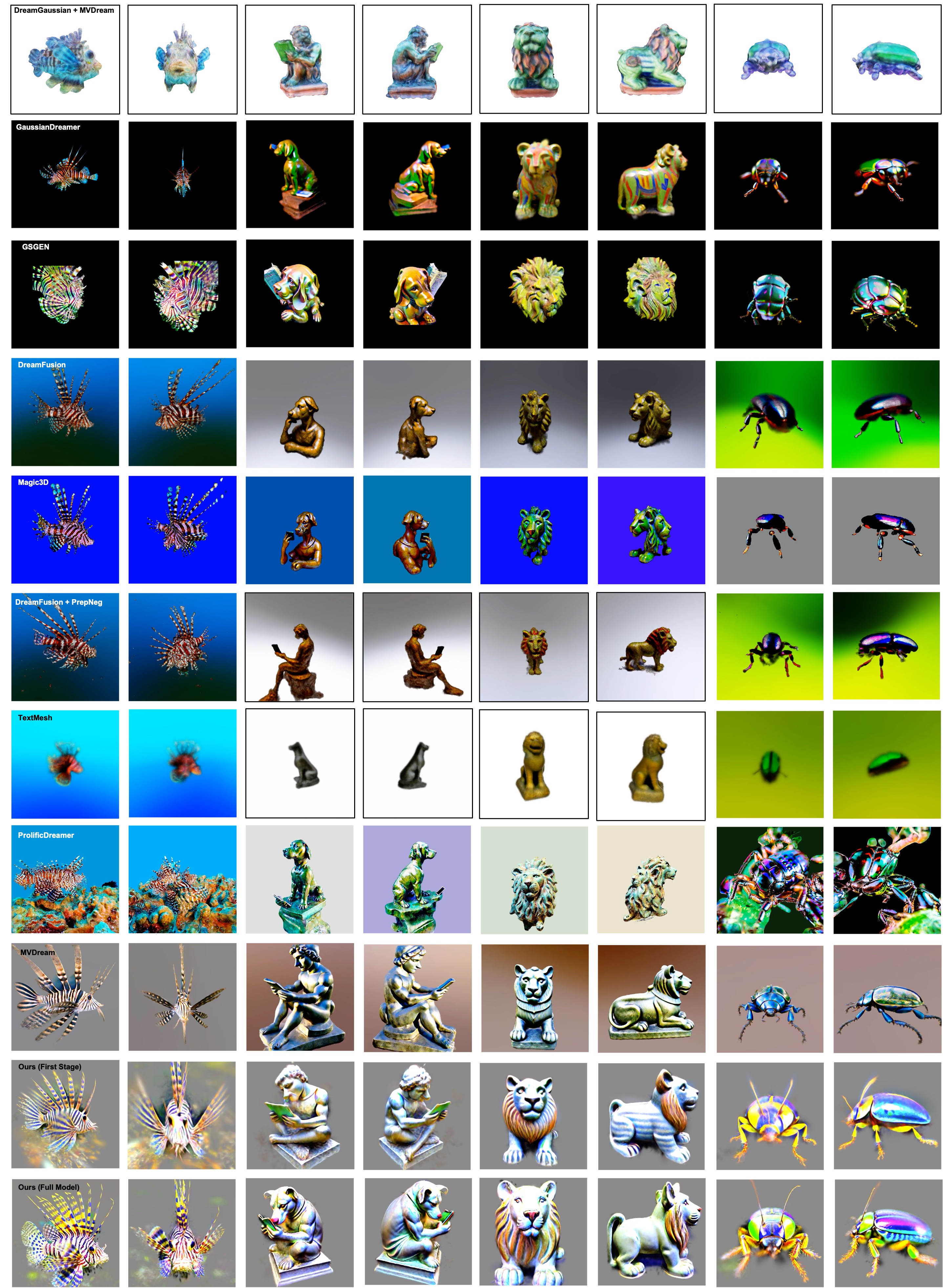}
    \caption{\footnotesize\textit{Qualitative comparison.} Each row shows two multi-view renderings of 3D models generated by baselines and our methods method using different prompts.
    Prompts used (left to right): ``a beautiful rainbow fish", ``Michelangelo style statue of dog reading news on a cellphone", ``a ceramic lion", ``a zoomed out DSLR photo of a shiny beetle"}
    \label{fig-supp_qualitative_0}
\end{figure*}

\begin{figure*}[h]
     \centering
    \includegraphics[width=0.93\linewidth]{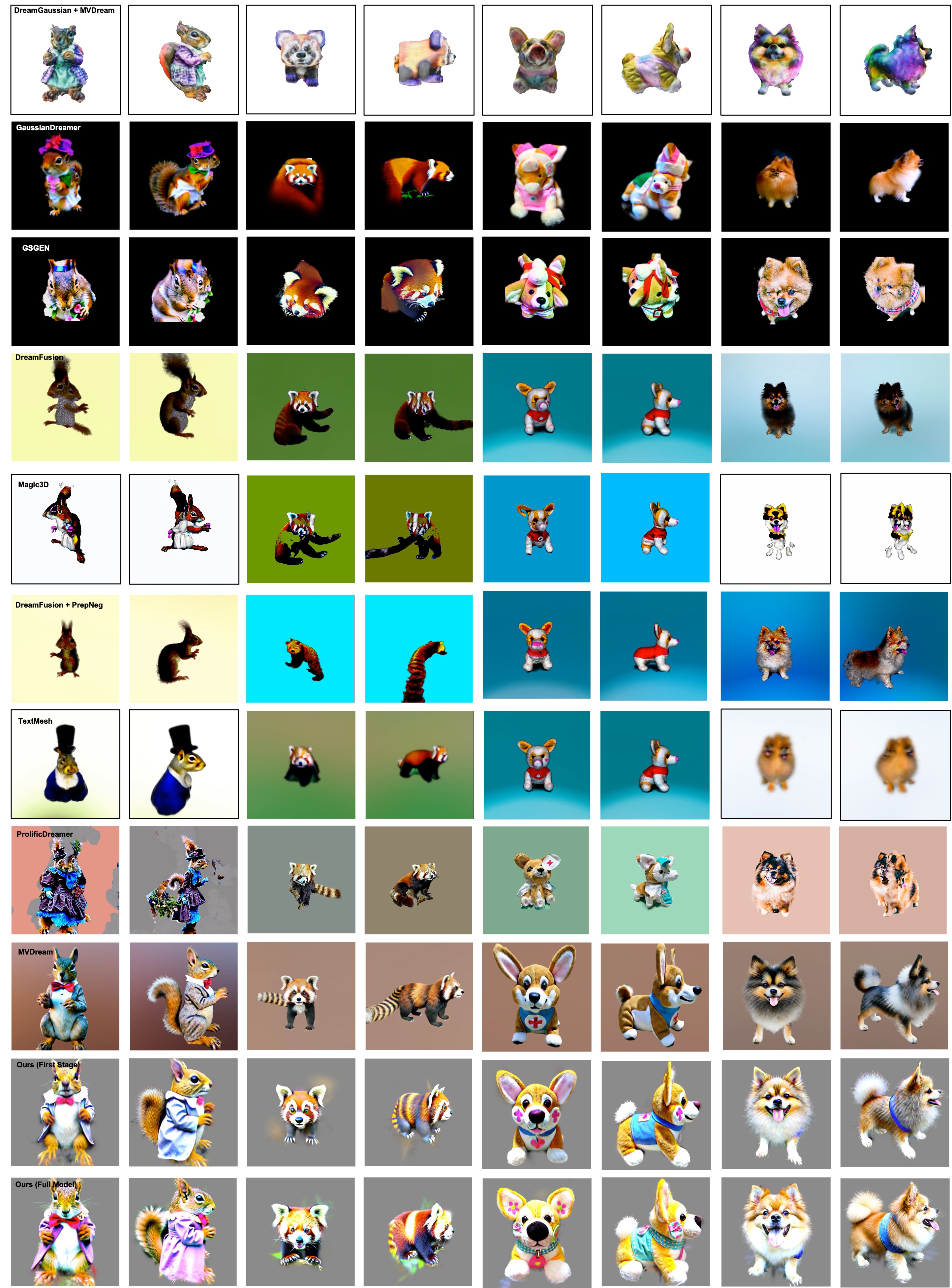}
    \caption{\footnotesize\textit{Qualitative comparison.} Each row shows two multi-view renderings of 3D models generated by baselines and our methods method using different prompts.
    Prompts used (left to right): ``a zoomed out DSLR photo of a squirrel dressed up like a Victorian woman", ``a red panda", ``a plush toy of a corgi nurse", ``a DSLR photo of a pomeranian dog"}
    \label{fig-supp_qualitative_1}
\end{figure*}

\begin{figure*}[h]
     \centering
    \includegraphics[width=0.93\linewidth]{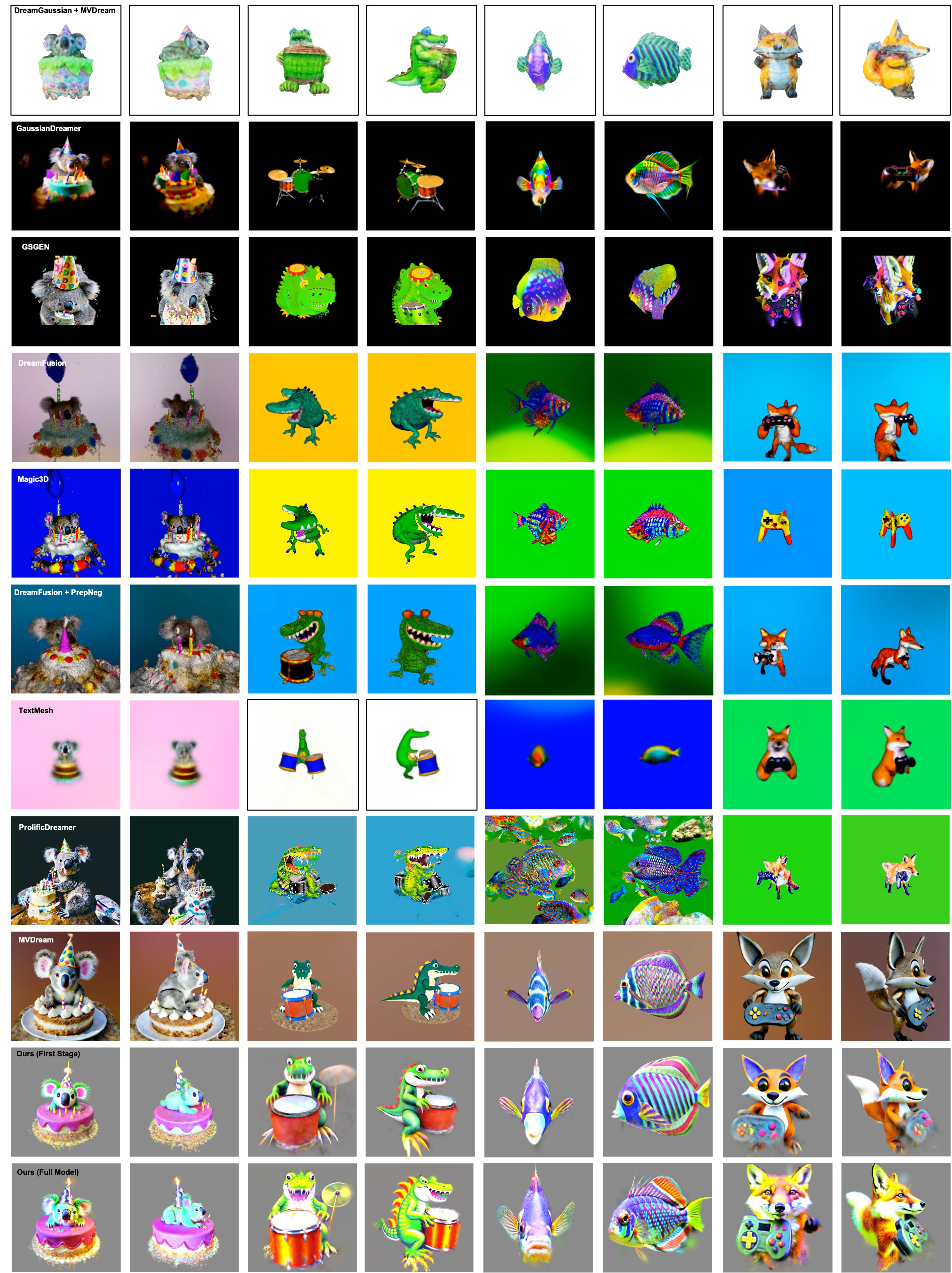}
    \caption{\footnotesize\textit{Qualitative comparison.} Each row shows two multi-view renderings of 3D models generated by baselines and our methods method using different prompts.
    Prompts used (left to right): ``a DSLR photo of a koala wearing a party hat and blowing out birthday candles on a cake", ``a crocodile playing a drum set", ``a lionfish", ``a DSLR photo of a fox holding a videogame controller"}
    \label{fig-supp_qualitative_2}
\end{figure*}

Figures~\ref{fig-supp_qualitative_0}, \ref{fig-supp_qualitative_1}, and \ref{fig-supp_qualitative_2} show additional qualitative comparisons of the state-of-the-art methods and the proposed approach.

\end{document}